\documentclass[letterpaper, 10 pt, conference]{ieeeconf}  % Comment this line out if you need a4paper

\IEEEoverridecommandlockouts                              % This command is only needed if 
\usepackage{amsmath} % assumes amsmath package installed
\usepackage{amssymb}  % assumes amsmath package installed
\usepackage{lipsum}
\usepackage[
backend=biber,
style=numeric,
sorting=none
]{biblatex}
\usepackage{xcolor}
\usepackage{changes}

\newcommand{\finnColor}{magenta!50}
\newcommand{\johannesColor}{cyan!50}
\newcommand{\erikColor}{red!50}
\newcommand{\stefanColor}{yellow!50}

\newcommand{\todoF}[1]{\todo[color=\finnColor]{#1}}
\newcommand{\todoJ}[1]{\todo[color=\johannesColor]{#1}}
\newcommand{\todoE}[1]{\todo[color=\erikColor]{#1}}
\newcommand{\todoS}[1]{\todo[color=\stefanColor]{#1}}

\usepackage{caption}
\usepackage{subcaption}

\presetkeys%
{todonotes}{inline,backgroundcolor=yellow}{disable}
\title{\LARGE \bf
Towards Interpretable Reinforcement Learning \\ with Constrained Normalizing Flow Policies
\todoJ{Towards...}
\todoF{I like it, added}
}

\author{Finn Rietz$^{1\dag}$, Erik Schaffernicht$^{1}$, Stefan Heinrich$^{2}$,  and Johannes A. Stork$^{1}$% <-this % stops a space
\thanks{$^{1}$Adaptive Interpretable Learning Systems, \"Orebro University, Sweden}
\thanks{$^{2}$IT University of Copenhagen, Denmark}
\thanks{$^\dag$Correspondance: finn.rietz@oru.se.}
\thanks{This work was partially supported by the Wallenberg AI, Autonomous Systems and Software Program (WASP) funded by the Knut and Alice Wallenberg Foundation.}
}

\begin{document}

\maketitle
\thispagestyle{empty}
\pagestyle{empty}

%%%%%%%%%%%%%%%%%%%%%%%%%%%%%%%%%%%%%%%%%%%%%%%%%%%%%%%%%%%%%%%%%%%%%%%%%%%%%%%%
\begin{abstract}
% \todo{Write one statement about the general problem} 
Reinforcement learning policies are typically represented by black-box neural networks, which are non-interpretable and not well-suited for safety-critical domains.
%\todo{Tell what this paper is about} 
To address both of these issues, we propose constrained normalizing flow policies as interpretable and safe-by-construction policy models.
%\todo{Describe how it solves the problem}
We achieve safety for reinforcement learning problems with instantaneous safety constraints, 
% for which we can construct the normalizing flow analytically.
%for which we can construct the normalizing flow's constituent transformations analytically.
for which we can exploit domain knowledge by analytically constructing a normalizing flow that ensures constraint satisfaction.
%\todo{Emphasize what is new or better}
The normalizing flow corresponds to an interpretable sequence of transformations on action samples,
% that ensures constraint satisfaction. 
each ensuring alignment with respect to a particular constraint.
%\todo{Mention the evidence indicating the advantages of the proposed approach} 
Our experiments reveal benefits beyond interpretability in an easier learning objective and maintained constraint satisfaction throughout the entire learning process. 
% With our approach we advocate for leveraging constraints over reward engineering, not only for improved interpretability and safety but also as a direct way of providing background knowledge to the agent that does not have to be learned through a complex reward signal.
Our approach leverages constraints over reward engineering while offering enhanced interpretability, safety, and direct means of providing domain knowledge to the agent without relying on complex reward functions.

\todoS{"uninterpreable" is a non-falsifiable term, can you be a bit softer? In any case, for the flow of reading, I would write "non-interpretable" in the first sentence}
\todoF{Done.}
\todoJ{Potential additional arguments:

-- Differentiable (for end-to-end RL) through multiple layers of constraints. The constraints are implemented in a differentiable way in the policy model.

-- Constraints are a better way to implement background knowledge as compared to reward. Rewards are messy and can lead to unexpected results (referring to the blog post).

-- Argue with efficiency: Constructed constraints as in this paper do not have to be learned as compared by using reward to model constraints.
}
\todoF{Tried to addresse the last two points with additional, last sentence in abstract.}
% \todoJ{Constituent transformations: unclear formulation.}
% changed sentence above, F.
\end{abstract}
%%%%%%%%%%%%%%%%%%%%%%%%%%%%%%%%%%%%%%%%%%%%%%%%%%%%%%%%%%%%%%%%%%%%%%%%%%%%%%%%
\section{INTRODUCTION}
\todoF{A comment by Finn.}
\todoJ{A comment by Johannes.}
\todoE{A comment by Erik.}
\todoS{A comment by Stefan.}
\todoS{Bring in the approximate nature of NNS, so in stead of "only converge to a safe agent" write "only gradually approximate a safe agent"?}

The trial-and-error nature of Reinforcement Learning (RL) algorithms and the black-box-like characteristic of monolithic neural network policies results in agents that are uninterpretable and poorly suited for safety-critical applications, especially those involving human participation. To account for the safety of RL agents, constrained RL methods~\cite{liu2021crlsurvey, achiam2017constrained, altman2021constrained, alshiekh2018safe} aim to obtain an agent that respects a set of (safety-) constraints. However, these methods typically require access to the transition dynamics of the environment or only obtain an approximately safe agent in the limit, that executes unsafe actions during training and exploration, to \textit{learn} which actions violate the constraints. 
\todoJ{They have to learn the constraint from interaction / discover the constraint.}
\todoF{Added above.}
Furthermore, constrained RL methods still acquire monolithic neural network policies that hinder verification and interpretation of the learned behaviour, despite these being crucial requirements in safety-critical domains and when interacting with humans.
Reward or task decomposition agents~\cite{russell2003q, juozapaitis2019explainable, rietz2023hierarchical}, on the other hand, are interpretable, since their modular structure allows for inspection and verification of separate components in the agent. 
\todoS{Add one sentence so explain in a nutshell "how"?}
\todoF{Not much more to say here than "you can look at the components". I emphasized this more.}

\begin{figure}
    \centering
    \begin{subfigure}[b]{0.15\textwidth}
        \centering
        \includegraphics[width=\textwidth]{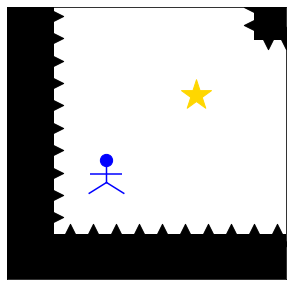}
    \end{subfigure}
    \begin{subfigure}[b]{0.15\textwidth}
        \centering
        \includegraphics[width=\textwidth]{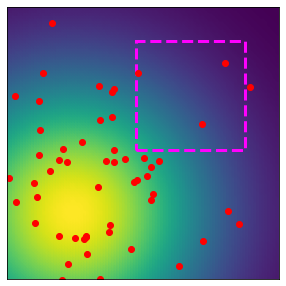}
    \end{subfigure}
    \begin{subfigure}[b]{0.15\textwidth}
        \centering
        \includegraphics[width=\textwidth]{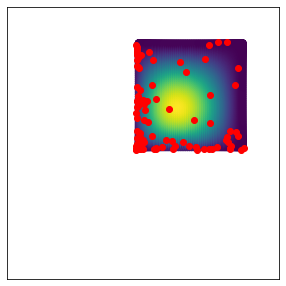}
    \end{subfigure}
    \caption{Our interpretable normalizing flow policy. \textbf{Left}: Environment, the agent should reach the star while avoiding dangerous obstacles and walls. \textbf{Middle}: A single flow step maps the initially unbounded policy distribution into the region satisfying the constraint (magenta rectangle), action samples are plotted in red. \textbf{Right}: The final policy distribution has support only over the allowed region.}
    \label{fig:interpretable_flow_policy}
\end{figure}

Therefore, to jointly increase the safety and interpretability of RL agents, we propose a modular and interpretable policy model that respects constraints even during learning and without requiring access to a complete model of the environment. 
Our method builds on recent normalizing flow policies~\cite{rezende2015variational, ward2019improving}, where a normalizing flow model is employed to learn a complex, multi-modal policy distribution. 
We show that by exploiting domain knowledge one can analytically construct intermediate flow steps that correspond to particular (safety-) constraints.
In such a setting, the flow-based policy is generated through an interpretable sequence of constraint-alignment steps. 
This is illustrated in Fig.~\ref{fig:interpretable_flow_policy} with only one constraint due to spatial constraints, examples with multiple constraints can be found on subsequent pages.
We refer to this model as a constrained normalizing flow policy (CNFP).
% CNFP allows for safe learning and exploration with multiple constraints and does not require a rejection-sampling step like most shielding approaches do, since the normalizing flow ensures that each action sample satisfies the constraints.
%
\todoJ{Mention that you can learn with all constraints present at the same time and that constraint violation does not have to be handled, e.g., by stopping the agent (as in Luc's paper). Assuming that you can stop the agent anywhere is not reasonable in real-world scenarios.}
\todoF{I need to see a bit first this is really the case for shielding. AFAIK some shields can be specified via LTL and I think they would forbid entering states in which the safety constraint (in the future) could not be satisfied.}
%

%\begin{figure}
%    \centering
%    \includegraphics[width=0.47\textwidth]{imgs/fig_1.png}
%     \caption{Our interpretable normalizing flow policy. \textbf{Left}: The agent should collect a coin while avoiding fire and water. \textbf{Middle}: A single flow step maps the initially unbounded policy distribution into the region satisfying the \textit{avoid fire} constraint (magenta ellipse). \textbf{Right}: The next flow step maps the density from the previous step into the \textit{avoid water} region (cyan ellipse).}
%    \label{fig:interpretable_flow_policy}
%\end{figure}

\section{Background}
We begin by formally defining the type of constrained RL problems we wish to solve and provide the relevant methods underlying our proposed approach.

\subsection{Constrained Reinforcement Learning}
Reinforcement learning problems are formalized as Markov Decision Processes (MDP)s. An MDP is a tuple $\mathcal{M} \equiv \langle \mathcal{S, A}, r, \rho, \gamma \rangle$, where $\mathcal{S}$ and $\mathcal{A}$ respectively denote the $n$- and $m$-dimensional state- and action-space, $r: \mathcal{S} \times \mathcal{A} \to \mathbb{R}$ is the scalar-valued reward function,
$\rho: \mathcal{S} \times \mathcal{A} \to \mathcal{S}$ are the discrete-time transition dynamics, and $\gamma \in [0, 1]$ is a discount factor. The goal in RL is to find a policy $\pi: \mathcal{S} \times \mathcal{A} \to [0, 1]$ that maximizes the total expected return
\begin{equation}
    J(\pi) =\mathbb{E}_{(\tau \sim \pi)} \Big[{\sum_{t=0}^\infty \gamma^t r(\mathbf{s}_t, \mathbf{a}_t)}\Big],
    \label{eq:return_objective}
\end{equation}
where $\mathbf{s}_t \in \mathcal{S}$ and $\mathbf{a}_t \in \mathcal{A}$ and $(\tau \sim \pi)$ is shorthand for denoting trajectories $\tau$ with actions sampled from the policy and states samples from the MDP's transition dynamics.
Constrained RL additionally assumes a number of constraints $c_1, \dots, c_K$ that limit policy search for Eq.~\eqref{eq:return_objective}. In this paper, we consider \textit{instantaneous constraints}~\cite{liu2021crlsurvey}, resulting in the constrained optimization
\begin{equation}
    \underset{\pi}{\max}\ J(\pi)\ \text{s.t.}\ c_k(\mathbf{s}_t, \mathbf{a}_t) \le \varepsilon_k\ \forall k \in \{1, \dots, K\}, \forall t,
    \label{eq:constrained_policy_search}
\end{equation}
where $\varepsilon_k$ is a pre-defined threshold for the constraint function $c_k: \mathcal{S} \times \mathcal{A} \to \mathbb{R}$.

A popular approach for solving such constrained RL problems relies on Lagrangian relaxation, which introduces Lagrange multipliers $\lambda$ to make for an approximation of the above constrained optimization. The Lagrangian relaxation of Eq.~\eqref{eq:constrained_policy_search} is given by
\begin{equation}
    J(\pi, \lambda) =\mathbb{E}_{(\tau \sim \pi)} \Big[
    \sum_{t=0}^\infty \gamma^t r(\mathbf{s}_t, \mathbf{a}_t)
    - \sum_{k=1}^K \lambda_k \big(c_k(\mathbf{s}_t, \mathbf{a}_t) - \varepsilon_k)\big)
    \Big].
    \label{eq:return_objective_lagrangian}
\end{equation}
Optimizing the objective in Eq.~\ref{eq:return_objective_lagrangian} with dual gradient descent, as in~\cite{ha2020learning, yang2021wcsac}, results in an agent that approximately solves Eq.~\eqref{eq:constrained_policy_search}. 
Other approaches to constrained RL involve projections, that learn to map actions into the allowed set~\cite{achiam2017constrained, pham2018optlayer}.
These methods often require an expensive optimization step, e.g. \citeauthor{pham2018optlayer}~\cite{pham2018optlayer} have to solve a quadratic problem to map each action into the safe set.
Shielding approaches~\cite{alshiekh2018safe, ijcai2023p637, hunt2021verifiably} ensure that only allowed actions are executed, however, they require access to the transition model or modify the environment directly to enforce constraints.
Importantly, solely rejecting actions that violate constraints does not suffice, since this would lead to biased gradient estimates~\cite{ijcai2023p637, chou2017improving}. 

In this paper, we instead exploit the following, useful property of instantaneous constraints, namely, the fact that they separate the per-state action space into two sub-spaces: $\mathcal{A}_{\varphi, k}^\mathbf{s}$, which contains all actions that satisfy constraint $k$ in state $\mathbf{s}$ and $\mathcal{A}_{\psi, k}^\mathbf{s}$, that contains the actions that violate constraint $k$ in state $\mathbf{s}$. We define $\mathcal{A}_{\varphi}^\mathbf{s} = \mathcal{A}_{\varphi, 1}^\mathbf{s} \cap \dots \cap \mathcal{A}_{\varphi, K}^\mathbf{s}$ as the intersection of all allowed constraint regions for state $\mathbf{s}$.
\todoJ{What if the sets are (for one constraint) are not contiguous (connected)? We could have several allowed and disallowed patches. This is currently not covered? Should this limitation be discussed?}
\todoF{I discuss this later. For now I am only considering convex constraints with one ``patch'' of allowed actions. There might not be an overlap between the patches of multiple constraints, which is a big problem for any constrained RL method, but ours handels this quite nicely, as discussed later.}
Therefore, instantaneous constraints can be used to induce a new MDP $\mathcal{M}_\varphi$~\cite{rietz2023prioritized, zhang2022lexicographic} that uses $\mathcal{A}_\varphi^\mathbf{s}$ as per-state action-space and leaves everything else as in the original MDP $\mathcal{M}$. Theoretically, $\mathcal{M}_\varphi$ can then be optimized with regular RL algorithms that then satisfy the constraints, even during learning, by construction. 
\todoJ{Should we discuss what you usually have to do when you hit a constraint? You have to stop the agent or re-map the agant to be safe with some background knowledge magic. Luc's paper argues that this is bad for on-policy learning. You will always be on a constraint compliant policy.}
\todoF{Yes, good point, I have extended the related work section above.}
In practice, this requires sample access to $\mathcal{A}_\varphi$ or a mapping from $\mathcal{A}$ to $\mathcal{A}_\varphi$. In this work we consider the former approach by exploiting mapping functions for instantaneous constraints and show how they can be integrated into Soft Actor-Critic (SAC) (or other policy-gradient algorithms) by means of normalizing flow policies.

\subsection{Soft Actor-Critic}
Soft Actor-Critic~\cite{haarnoja2018soft, haarnoja2018softAA} is a model-free RL algorithm for MDPs with continuous state and action spaces. SAC maximizes the following maximum-entropy objective~\cite{ziebart2008maximum}, which augments Eq.~\eqref{eq:return_objective} with the policy's entropy $\mathcal{H} = \mathbb{E}_{\mathbf{a} \sim \pi(\cdot  \mid \mathbf{s})}[- \log \pi(\mathbf{a} \mid \mathbf{s})]$,
\begin{equation}
    J_\text{ME}(\pi) =\mathbb{E}_{(\mathbf{a}_t \sim \pi), (\mathbf{s}_t \sim \rho)} \Big[{\sum_{t=0}^\infty \gamma^t r(\mathbf{s}_t, \mathbf{a}_t) + \alpha \mathcal{H}(\pi(\cdot \mid \mathbf{s}_t)}\Big],
    \label{eq:me_objective}
\end{equation}
where $\alpha$ balances the entropy and the reward objective.
SAC learns an on-policy critic Q-function, $Q_\theta$, with parameter $\theta$, by optimizing for Bellman consistency
\begin{equation}
    J_Q(\theta) = \mathbb{E}_{\mathcal{D}} 
    \Big[
    \frac{1}{2}
    \Big(
    Q_\theta(\mathbf{s}_t, \mathbf{a}_t) - 
    \big(
    r(\mathbf{s}_t, \mathbf{a}_t) + \gamma V_{\bar{\theta}} (\mathbf{s}_{t+1})
    \big)
    \Big)^2
    \Big
    ],
    \label{eq:q_bellman_backup}
\end{equation}
where $\mathbf{s}_t, \mathbf{a}_t$, and $\mathbf{s}_{t+1}$ are sampled from a replay buffer $\mathcal{D}$, $V_{\bar{\theta}} = \mathbb{E}_{(\mathbf{a}_{t+1} \sim \pi})[Q_{\bar{\theta}}(\mathbf{s}_{t+1}, \mathbf{a}_{t+1}) - \alpha \log \pi(\mathbf{a}_t \mid \mathbf{s}_t))]$ is the maximum-entropy on-policy state-value function, and $\bar{\theta}$ is a target network~\cite{mnih2015human} parameter.
With respect to the actor, SAC employs an infinite-support, unimodal Gaussian with diagonal covariance and mean given by a policy network, $\pi_\phi$, which is parameterized by $\phi$. The policy network update makes use of the \textit{reparametrization trick} and backpropagates through the critic
\begin{equation}
    J_\pi(\phi) = \mathbb{E}_{(\mathbf{s}_t \sim \mathcal{D})}
    \Big[
    \mathbb{E}_{(\mathbf{a}_t \sim \pi_\phi)}
    \big[
    \alpha \log \pi_\phi(\mathbf{a}_t \mid \mathbf{s}_t) - Q_\theta(\mathbf{s}_t, \mathbf{a}_t)
    \big]
    \Big],
    \label{eq:policy_objective}
\end{equation}
to increase the likelihood of actions that have high Q-values.
\todoJ{We should highlight that this is a practical solution that is not part of the main paper. They do that to make it work on their robot / environment.}
\todoF{I don't follow. The above is just the SAC policy update from the original paper. Should I not mention that it is relying on reparametrization / stochastic gradients?}
To bound the action space, SAC \textit{squashes} the Gaussian action samples with the hyperbolic tangent function to obtain the final actor distribution, which is referred to as a squashed Gaussian distribution. 
The density of the squashed Gaussian can be obtained using the change of variables formula.
Given a random variable $\mathbf{a}$, its density $\pi(\mathbf{a} \mid \mathbf{s})$, and an invertible function $f$, the density of the transformed random variable $\mathbf{a'} = f(\mathbf{a})$ is given by
\begin{equation}
    \pi(\mathbf{a'} \mid \mathbf{s}) 
    = 
    \pi(\mathbf{a} \mid \mathbf{s}) \left| \det \frac{\partial f^{-1}}{\partial \mathbf{a'}} \right|
    = 
    \pi(\mathbf{a} \mid \mathbf{s}) \left| \det \frac{\partial f}{\partial \mathbf{a}} \right|^{-1},
    \label{eq:change_of_variables}
\end{equation}
where $f^{-1}$ is the inverse of the transformation $f$. Eq.~\eqref{eq:change_of_variables} is used to obtain the policy's log-density in Eq.~\eqref{eq:q_bellman_backup} and \eqref{eq:policy_objective}.
Importantly, while SAC uses the hyperbolic tangent for $f$ to bound the action space, the change of variables formula allows for \textit{any} invertible function. This prompts the key idea behind our method: If we can express instantaneous constraints in terms of invertible functions on $\mathcal{A}$, we can directly transform $\mathcal{A}$ into $\mathcal{A}_\varphi$ on a per-state basis and learn optimal policies directly in $\mathcal{M}_\varphi$. 
In the next section, we show how this idea can be generalized to multiple constraints and that the resulting distribution corresponds exactly to what is known as a \textit{normalizing flow policy}.

\section{Constrained Normalizing Flow Policies}
Normalizing Flows (NFs)~\cite{rezende2015variational} are models for variational inference that transform a simple, initial density (e.g. Gaussians) into a complex posterior distribution by applying a sequence of learned, invertible transformations to the original density.
A degenerate, one-step NF with only one transformation $f$ is referred to as a \textit{flow} and the resulting density is given by Eq.~\eqref{eq:change_of_variables}.
In this sense, unmodified SAC applies a one-step NF to obtain the density of the squashed Gaussian distribution, which has no parametric form.
A proper NF refers to the composition of multiple (learned) transformations on the random variable $\mathbf{a}$, with $\mathbf{a}_M = f_M(f_{M-1}(\dots f(\mathbf{a}))$, in which case Eq.~\eqref{eq:change_of_variables} is successively applied to yield the (log-) density
\begin{equation}
    \log \pi(\mathbf{a}_M \mid \mathbf{s}) = \log \pi(\mathbf{a} \mid \mathbf{s}) - \sum_{m=1}^M \log \left| \det \frac{\partial f_m}{\partial \mathbf{a}_{m-1}} \right|.
    \label{eq:normalizing_flow}
\end{equation}
As shown in~\cite{rezende2015variational} NFs are highly flexible and can approximate complex, multi-modal posterior distributions. By viewing the squashed Gaussian distribution in SAC as the result of single flow step, it is intuitive to replace the squashed Gaussian with more expressive, multi-modal NFs. When these transformations are learned, the resulting architecture is referred to as a \textit{normalizing flow policy}~\cite{ward2019improving}.

Normalizing flow policies are primarily used to obtain more expressive policy distributions, which reportedly improves exploration and learning efficiency~\cite{ward2019improving, delalleau2019discrete, mazoure2020leveraging}.
We find two related works investigating constrained RL problems with the help of flow-based policies.
\citeauthor{brahmanage2024flowpg}~\cite{brahmanage2024flowpg} also focus on RL problems with instantaneous constraints, as in Eq.~\eqref{eq:constrained_policy_search}, and propose \textit{FlowPG} to learn an invertible mapping from $\mathcal{A}$ to $\mathcal{A}_\varphi$, i.e. to map any given action into the per-state allowed sub-space. FlowPG requires a large dataset of actions in $\mathcal{A}_\varphi^\mathbf{s}$ for each state $\mathbf{s}$, which the authors propose to obtain via an expensive, initial Hamiltonian Monte-Carlo (HMC)~\cite{betancourt2017conceptual} step. 
\citeauthor{chen2023generative}~\cite{chen2023generative} instead focus on feasibility constraints in large, discrete action spaces and learn an \textit{Argmax Flow}~\cite{hoogeboom2021argmax} network that generates samples from the feasible, categorical action distribution. To ensure that only allowed actions are executed, the authors include a rejection sampling step that may fail if the feasible set is small.
The key difference % to our method 
between previous works on constrained flow-based policies and our method
is that we, instead of learning the $M$ transformations from data, propose to construct the transformation functions analytically, considering domain knowledge and constraint functions. We describe our approach to this in the next section.

\todoJ{From reading this, it is unclear to me whether they (SOTA) are doing constraints or not.}
\todoF{The main use of flow-based policy is to boost exploration through complex NF distributions. I found two works that use NFs for constrained RL. They learn NFs that map into the constrain regions. I tried to clarify this above.}

\section{Method}
\begin{figure}
     \centering
     \begin{subfigure}[b]{0.23\textwidth}
         \centering
         \includegraphics[width=\textwidth]{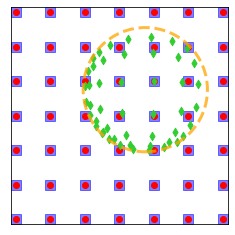}
     \end{subfigure}
     \hfill
     \begin{subfigure}[b]{0.23\textwidth}
         \centering
         \includegraphics[width=\textwidth]{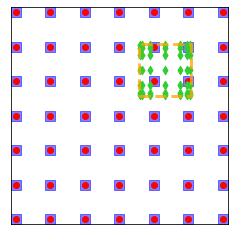}
     \end{subfigure}
    \caption{Invertible mapping functions. Exemplary constraint regions are drawn in orange \includegraphics[height=0.6\baselineskip]{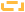}. These functions map from the unbounded domain into the constraint region, i.e. $f(\includegraphics[height=0.6\baselineskip]{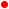}) \to \includegraphics[height=0.6\baselineskip]{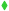}$. The inverse function maps from the constraint region back to the unbounded domain, i.e. $f^{-1}(\includegraphics[height=0.6\baselineskip]{imgs/squashed_particle.png}) \to \includegraphics[height=0.6\baselineskip]{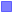} = \includegraphics[height=0.6\baselineskip]{imgs/particle.png} $.}
    \label{fig:exemplary_mappings}
\end{figure}

In this paper, we consider RL problems with instantaneous constraints, as in Eq.~\eqref{eq:constrained_policy_search}, with the constraints defined on the MDPs state-action space. We propose to analytically construct functions that map into the per-state, per-constrain allowed action subspace $\mathcal{A}_{\varphi, k}^\mathbf{s}$. More formally, we assume $K$ instantaneous constraint functions $c_1(\mathbf{s,a}), \dots, c_K(\mathbf{s, a})$ with corresponding thresholds $\varepsilon_1, \dots, \varepsilon_K$, which induce $\mathcal{A}_{\varphi, 1}^\mathbf{s}, \dots, \mathcal{A}_{\varphi, K}^\mathbf{s}$, the per-state $\mathbf{s}$, per-constrain $k$ allowed action sub-spaces.
We now make a relatively strong, simplifying assumption: We assume that all $\mathcal{A}_{\varphi, k}^\mathbf{s}$ are convex because it is relatively easy to find invertible transformations that map points into the convex sets. So far, we developed functions for mapping or \textit{squashing} into hypercubes, hyperspheres, and ellipsoids (the former could correspond to percentiles on learned Gaussians), but we hypothesize that there are invertible squashing functions for more complex polytopes. Fig.~\ref{fig:exemplary_mappings} illustrates these invertible squashing functions in 2D. 
While the exploration of non-convex, invertible squashing functions is out of scope for this paper, we note that multiple simple, convex constraints can be combined to induce a complex, overall constraint on the agent.

\todoJ{You can argue that in control (e.g., the STESS paper) everything is build from simpler, convex constraints / tasks. You can combine many convex constraints to make a complex constraint.}
\todoF{I don't know STESS, but its a good points regardless. Added above!}

\begin{figure}
    \centering
    \includegraphics[width=0.5\textwidth]{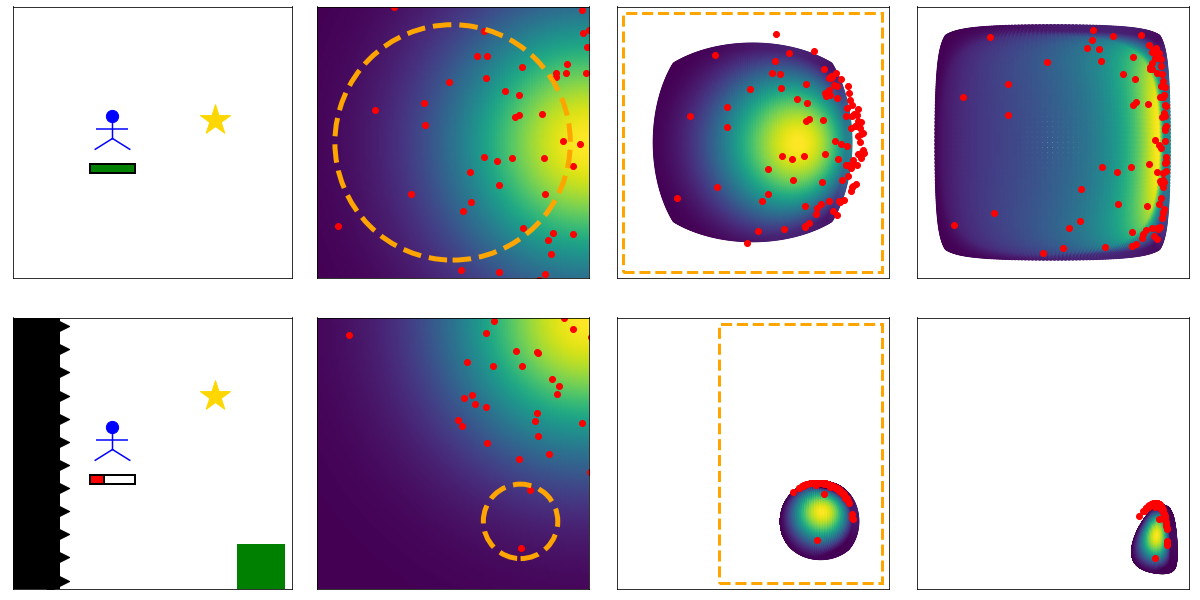}
    \caption{Two normalizing flows. \textbf{Top}: Permissive constraints, no obstacles are in close proximity and the battery is fully charged. \textbf{Bottom}: Restrictive constraints, the agent is close to an obstacle and its battery is almost empty. The green rectangle indicates a charging station.}
    \label{fig:large-vs-small-constraints}
\end{figure}

To provide a concrete example of our method, consider two constraint function $c_1(\mathbf{s, a})$ and $c_2(\mathbf{s, a})$ with corresponding $\varepsilon_1, \varepsilon_2$, whose convex sub-spaces are $\mathcal{A}_{\varphi, 1}^\mathbf{s}$ and $\mathcal{A}_{\varphi, 2}^\mathbf{s}$ in each state $\mathbf{s}$, and the corresponding invertible functions $f_1^\mathbf{s}$ and $f_2^\mathbf{s}$ that respectively map points into $\mathcal{A}_{\varphi, 1}^\mathbf{s}$ and $\mathcal{A}_{\varphi, 2}^\mathbf{s}$.
By assuming that $f_1^\mathbf{s}$ and $f_2^\mathbf{s}$ are known (or can be constructed), we can directly use these functions to construct an interpretable, constrained NF policy by inserting them into Eq.~\eqref{eq:normalizing_flow}.
The resulting NF policy will sequentially map the initially unbounded action-space into the sub-space allowed by the constraints, such that the final distribution only has support over $\mathcal{A}_{\varphi, 1}^\mathbf{s} \cap \mathcal{A}_{\varphi, 2}^\mathbf{s}$, the intersection of all constraint subsets. This is the core idea behind our method, which is applicable to instantaneous constraints for which we know or can analytically construct the mapping functions. 
The consequences of this modelling can be seen in Fig.~\ref{fig:large-vs-small-constraints}. As shown, depending on the state and the constraint, the squashing functions are restrictive or largely permissive. The final policy distribution is obtained in an interpretable manner since each transformation in the normalizing flow aligns the policy with respect to one of the 
constraints.
This approach can trivially be extended to $K > 2$ constraints, since simply the range of the summation in Eq.~\ref{eq:normalizing_flow} increases from $2$ to $K$.

Note, that the order in which we apply our transformations matters since the mapping functions are or can be non-linear, meaning we have $f_1^\mathbf{s}(f_2^\mathbf{s}(\mathbf{a})) \ne f_2^\mathbf{s}(f_1^\mathbf{s}(\mathbf{a}))$.
% The normalizing flow in Eq.~\eqref{eq:normalizing_flow} applies transformations sequentially, and since our squashing functions may be non-linear, order matters and we have $f_1^\mathbf{s}(f_2^\mathbf{s}(\mathbf{a})) \ne f_2^\mathbf{s}(f_1^\mathbf{s}(\mathbf{a}))$.
% For the an exemplary normaling flow like $f_2^\mathbf{s}(f_1^\mathbf{s}(\mathbf{a}))$, this means that $\mathcal{A}_{\varphi, 1}^\mathbf{s} = f_1^\mathbf{s}(\mathbf{a})$ would be transformed by $f_2^\mathbf{s}$ and projected into $\mathcal{A}_{\varphi, 2}^\mathbf{s}$. 
The normalizing flow generally maps into the intersection of $\mathcal{A}_{\varphi, 1}^\mathbf{s}$ and $\mathcal{A}_{\varphi, 2}^\mathbf{s}$, however, the resulting distribution can look different, depending on the order of transformations. 
This can be used to impose a notion of priority on the constraints: For each constraint $l$ and the corresponding transformation $f_l$ that comes before $k$ in the flow sequence, the intermediate $\mathcal{A}_{\varphi, l}^\mathbf{s}$  will be mapped into $\mathcal{A}_{\varphi, k}^\mathbf{s}$ by the subsequent transformation $f_k$ which may not overlap with $\mathcal{A}_{\varphi, l}^\mathbf{s}$. This is a desirable property because it ensures that even if constraints are incompatible, i.e. when $\mathcal{A}_{\varphi, 1}^\mathbf{s} \cap \mathcal{A}_{\varphi, 2}^\mathbf{s} = \emptyset$, our NF policy will always map into the sub-space of the higher-priority constraint, whose transformation is applied after that of the lower-priority constraints.

\begin{figure*}
     \centering
     \includegraphics[width=0.6\textwidth]{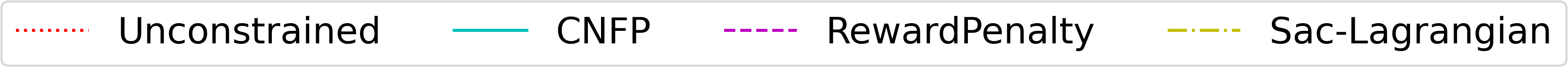}
     \vfill
     \hfill
     \begin{subfigure}[b]{0.49\textwidth}
         \centering
         \includegraphics[width=\textwidth]{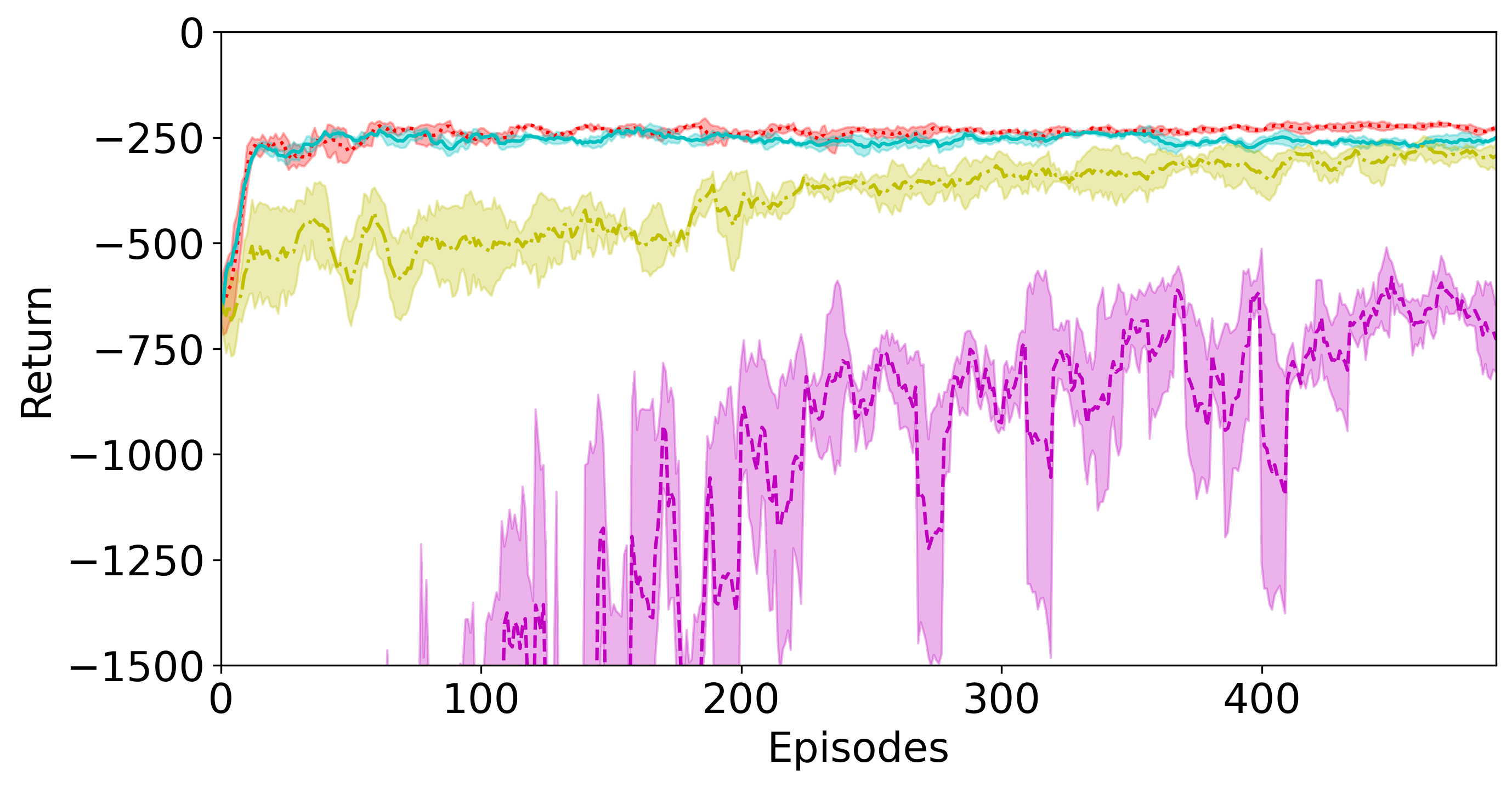}
     \end{subfigure}
     \hfill
     \begin{subfigure}[b]{0.49\textwidth}
         \centering
         \includegraphics[width=\textwidth]{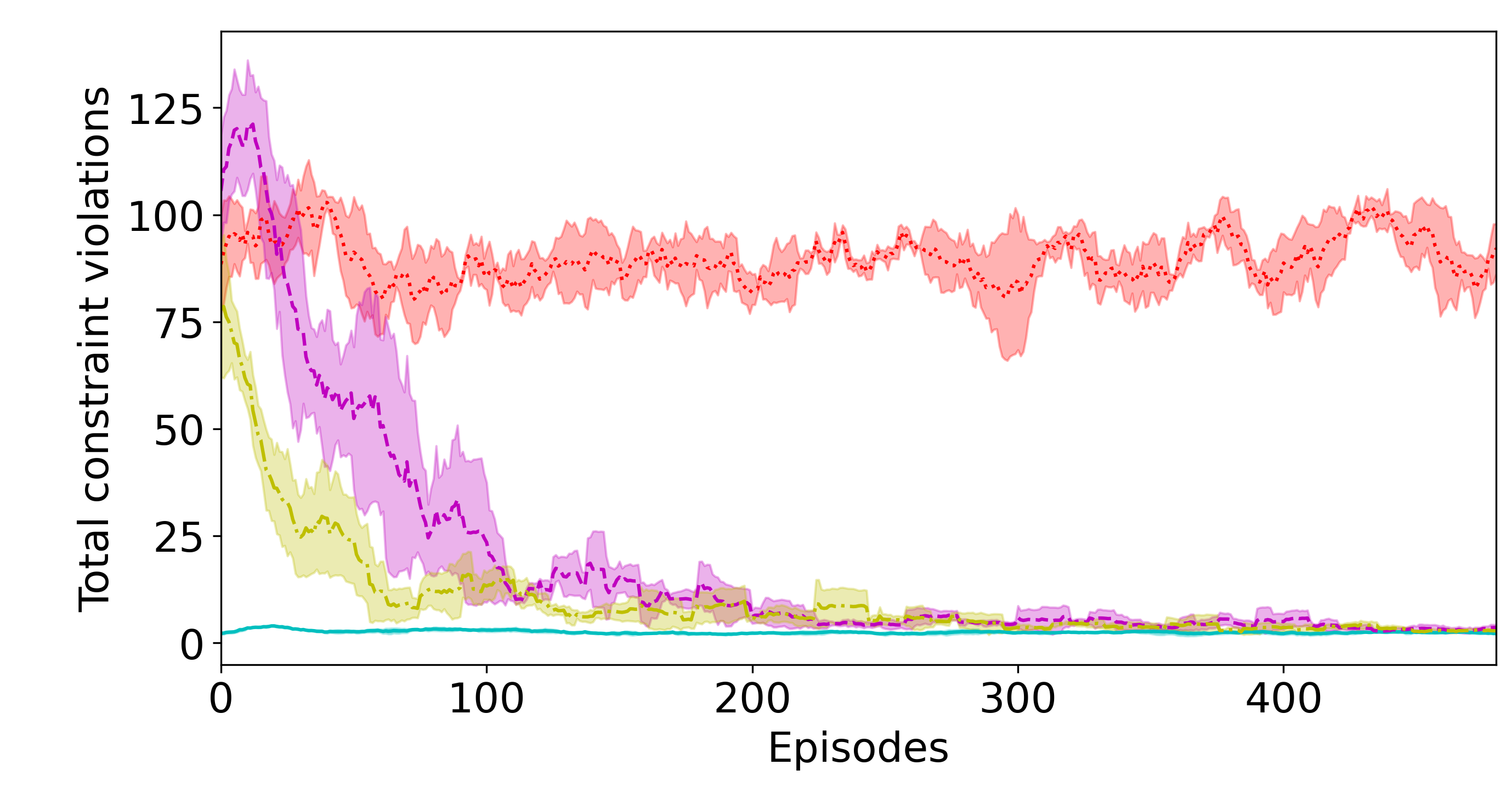}
     \end{subfigure}
     \hfill
        \caption{Baseline comparison in a constrained 2D point navigation environment. \textbf{Left}: 
        % Our agent (CNFP) learns the task quickly due to a smoother reward function and smaller search space. 
        Our agent (CNFP) learns the task as quickly as the unconstrained agent since it optimizes the same, smooth and dense reward function while benefiting from a reduced search space.
        \textbf{Right}: 
        % Our agent maintains near-optimal constraint-satisfaction throughout learning. 
        Unlike other baselines, our agent maintains quasi-perfect constraint satisfaction throughout learning.
        The experiment was repeated three times with varying random seeds, the shaded area corresponds to one standard deviation around the mean.}
        \label{fig:constraint-return-result}
\end{figure*}

The concrete steps of our method can be summarized as follows: Given an RL problem with a set of instantaneous, convex constraints $c_1(\mathbf{s, a}) \le \varepsilon_1, \dots, c_K(\mathbf{s, a}) \le \varepsilon_K$, find the corresponding invertible functions $f_1, \dots, f_K$ that map into $\mathcal{A}_{\varphi, 1}, \dots \mathcal{A}_{\varphi, K}$. Next, order the constraints by domain-specific priority, e.g. $c_1 \succ \dots \succ c_K$. Insert the constraint mapping function into Eq.~\eqref{eq:normalizing_flow}, such that the lowest-priority constraint transformation $f_K$ is applied first and the highest priority constraint transformation, $f_1$ is applied last. Use the resulting normalizing flow to compute the log-density of the constrained NF policy, e.g. in Eq.~\eqref{eq:q_bellman_backup} and Eq.~\eqref{eq:policy_objective} for SAC or other policy-gradient algorithms.
In the next section, we demonstrate this approach in a simplistic 2D environment as proof-of-concept.

\section{Experiments}
\subsection{Environment}
We empirically validate our method on a constrained 2D point navigation problem, where the agent has to reach a target coordinate while constrained to avoiding obstacles and keeping the battery charge above 20\%. At each step, the battery depletes by 1\% but can be charged by visiting a charging station, with one charging station placed at each side of the rectangular environment. The outside walls as well as a centrally-placed rectangle provide the static obstacles the agent should avoid colliding with. 
The central obstacle ensures that the direct path to the goal is obstructed, such that an unconstrained agent will inquire high constraint violations.
The observation vector is in $\mathbb{R}^5$ and corresponds to the agent's current 2D coordinate, the current battery level, and the 2D goal coordinate. Actions correspond to translations in the 2D plane. 
The reward function is dense and corresponds to the negative Euclidean distance between the agent and the target coordinate, encouraging the agent to greedily navigate towards the goal. When the agent reaches the goal, a bonus of $10$ reward is provided and a new goal position is randomly sampled. There are no terminal states, however, episodes are truncated and the environment resets after 100 steps.
The constraints are modelled as indicator functions. The constraint for obstacle avoidance, $\mathbb{I}_O(\mathbf{s,a})$ maps to one if executing $\mathbf{a}$ in $\mathbf{s}$ would lead to a collision with an obstacle. The constraint function for the battery level, $\mathbb{I}_B(\mathbf{s,a})$ maps to one if executing $\mathbf{a}$ in $\mathbf{s}$ leads to the battery falling beneath the 20\% threshold. Both corresponding constraint thresholds, $\varepsilon_O, \varepsilon_B$ are set to $0$.
 
\subsection{Methods}
For our constrained normalizing flow policy (CNFP), we analytically construct the invertible functions $f_B^\mathbf{s}, f_O^\mathbf{s}$  for mapping into $\mathcal{A}_{\varphi, B}^\mathbf{s}$ and $\mathcal{A}_{\varphi, O}^\mathbf{s}$, i.e., the per-state action-subspaces that keep the battery and obstacle-avoidance constraints satisfied. Due to the rectangular layout of the environment we model $f_O^\mathbf{s}$, the function for mapping into $\mathcal{A}_{\varphi, O}^\mathbf{s}$, with the rectangle squashing function (right side of Fig.~\ref{fig:exemplary_mappings}). The dimensions of the rectangular constraint region are inferred from the environment and the agent's current position, just like $\mathbb{I}_O(\mathbf{s,a})$ itself. If the agent is sufficiently far from all obstacles, this constraint simply bounds the action space to $[-1, 1]$, however, in closer proximity to the obstacle, the bound becomes tighter and excludes those actions that would lead to collisions.
The battery constraint mapping function $f_B^\mathbf{s}$ is modelled with the circular squashing function (left side of Fig.~\ref{fig:exemplary_mappings}). If the battery level is sufficiently high, the circle is zero-centred and has a large radius. As the battery level decreases, the radius becomes smaller and the circle is placed at the closest charging station. Thus, if the battery level is low, the agent is automatically ``pulled'' to the closest charging station. We define the priority order of these constraints as $\mathbb{I}_O \succ \mathbb{I}_B$, meaning avoiding obstacles is assigned higher priority than keeping the battery fully charged. Our constrained normalizing flow policy thus corresponds to $\mathbf{a'} = f_O^\mathbf{s}(f_B^\mathbf{s}(\mathbf{a} \sim \pi(\cdot \mid \mathbf{s})))$.

In addition to our CNFP agent, we include the following baselines. Firstly, we include an unconstrained SAC agent that maximizes the reward function while disregarding the constraints entirely. This agent provides a baseline for constraint violations caused when optimizing only the reward.
Next, we include a SAC agent where constraint violations are punished via the reward function. For this agent, we modify the dense reward function to yield a large negative penalty, $-100$, whenever at least one of the two constraints is violated. Behaving optimal with respect to this reward function is equivalent to solving the task while respecting the constraints.
Lastly, we include a Sac-Lagrangian agent that optimizes Eq.~\ref{eq:return_objective_lagrangian}, as described in~\cite{ha2020learning, yang2021wcsac}.

\subsection{Results}
Our main result is shown in Fig.~\ref{fig:constraint-return-result}, with two main insights. 
Firstly, our CNFP agent learns to solve the goal-navigation task optimally within only a few episodes, as can be seen on the left side in Fig.~\ref{fig:constraint-return-result}.
This can be explained by two observations. 
%On the one hand, the reward function our CNFP agent is optimizing is smooth and dense, which makes learning easy.
On the one hand, CNFP optimizes the same smooth and dense reward function as the unconstrained baseline agent, which makes learning easy.
This is not the case for the reward penalty or Lagrangian agents. 
Although the Lagrangian agent also optimizes the same smooth and dense reward function as the unconstrained baseline and CNFP, it also has to consider the constraint violations in the actor objective, which results in a harder objective for policy search. 
As a consequence, the Lagrangian agent converges only slowly to near-optimal performance.
For the reward penalty agent, the reward function is still dense but it is not smooth since it is dominated by the large penalties raised due to constraint violations. 
This makes learning the task harder. 
The reward penalty agent did therefore only learn to avoid constraint-violating actions, but did not converge to an optimal level of performance within the number of episodes of this experiment. 
The quick convergence of our CNFP agent to an optimal level of task return can further be explained by a reduced search space. 
The behaviors accounting for the obstacle avoidance and battery-level constraints do not have to be learned, 
%since they are already encoded into the agent via the transformation functions $f_O^\mathbf{s}$ and $f_B^\mathbf{s}$. 
since we can exploit domain knowledge to encode them into the agent via the transformation functions $f_O^\mathbf{s}$ and $f_B^\mathbf{s}$. 
No matter where the policy network places the initial Gaussian $\mathcal{N}(\mu_\phi, \Sigma_\phi)$ and which action is sampled, if the agent is close to an obstacle, the sampled action (and corresponding density) will be transformed by $f_O^\mathbf{s}$ in such a way that a collision is no longer possible (e.g. Fig.~\ref{fig:interpretable_flow_policy} and Fig~\ref{fig:large-vs-small-constraints}). 
The same is true for the battery constraint.
Therefore, our agent's learning objective is solely the maximization of task return, while the constraints are satisfied by construction of the constrained normalizing flow policy and do not require any plasticity in the policy network.

The second insight is that, as can be seen on the right side of Fig.~\ref{fig:constraint-return-result}, our CNFP agent maintains quasi-perfect constraint satisfaction throughout the entirety of training. 
This is expected, since correct mapping functions $f_O^\mathbf{s}$ and $f_B^\mathbf{s}$ should never allow constraint-violating actions to be executed.
% The result of the CNFP agent is in stark contrast to the other methods.
% In contrast, all other agents inquire a high number of constraint violations.
As revealed through the unconstrained agent, solving the task optimally without accounting for constraints results in many constraint-violating actions.
Both the reward penalty and the Lagrangian agent drastically decrease the number of constraint violations throughout training, however, both inquire a high number of violations at the beginning of training. 
At the same time, for these agents, the reduction in constraint violations comes at a cost: The reward penalty agent learns overly pessimistic behaviour which indeed results in few violations, however, at the cost of drastic performance degradation in terms of task return.
The Lagrangian agent converges to a similar level of constraint violation as the reward penalty agent while achieving better task return, 
% although still considerably worse than the unconstrained and CNFP agents.
although only reaching near-optimal performance at the very end of training.
Thus, to summarize the results so far, our CNFP agent is the only one to maintain perfect constraint satisfaction through training while achieving task return levels as quickly as and on par with an unconstrained, optimal agent.

Lastly, we want to highlight again the interpretable nature of our model. 
Unlike all other baselines that learn a single, monolithic policy, our CNFP policy is interpretable, since each step in the normalizing flow can be visualized and used to explain the agent's behaviour. While the initial, unbounded policy distribution is still obtained from a black-box neural network, it can be explained how this distribution is transformed to ensure that the agent respects the given constraints. This can reveal flaws in the construction of the constraint mapping functions, i.e. it can be seen when a mapping function is to permissive and allows the execution of unsafe actions. It can therefore be explained how an unsafe action in particular state must be transformed to ensure alignment w.r.t each constraint.
This is not the case for our baseline methods that learn monolithic policies with constraints simply moved into the learning objective.

\section{Conclusion and future work}
In this paper, we have shown how normalizing flows can be used to obtain interpretable policies for constrained reinforcement learning problems.
Our experiments revealed a favourable comparison against baselines with respect to task return and constraint violations, with additional benefits in and beyond interpretability: The action-space transformation functions, which constitute the normalizing flow, make for a form of knowledge transfer by encoding desired behaviour in terms of constraints on the action space. Then, the constraints do not have to be considered in the reward function and value estimates, which leaves a simple optimization objective that can be learned quickly.
As the most important future work, we see the development of non-convex transformation functions, to broaden the applicability of our approach to more complex scenarios, as well as the integration of learnable mapping functions for complex constraints.
In this context, it might be worthwhile to explore differentiable constraint functions, as in ~\cite{ijcai2023p637}, which are susceptible to learning with normalizing flows.

\todoJ{From recent reviewing, I think we should think about using this in offline-to-online RL and continuous learning. The constraints could change over time and we would have to change the policy over time and stop it from forgetting, which is very interesting.}

\addtolength{\textheight}{-1cm}   % This command serves to balance the column lengths

%%%%%%%%%%%%%%%%%%%%%%%%%%%%%%%%%%%%%%%%%%%%%%%%%%%%%%%%%%%%%%%%%%%%%%%%%%%%%%%%

%%%%%%%%%%%%%%%%%%%%%%%%%%%%%%%%%%%%%%%%%%%%%%%%%%%%%%%%%%%%%%%%%%%%%%%%%%%%%%%%

%%%%%%%%%%%%%%%%%%%%%%%%%%%%%%%%%%%%%%%%%%%%%%%%%%%%%%%%%%%%%%%%%%%%%%%%%%%%%%%%
% \section*{ACKNOWLEDGMENT}
% This work was partially supported by the Wallenberg AI, Autonomous Systems and Software Program (WASP) funded by the Knut and Alice Wallenberg Foundation.
%%%%%%%%%%%%%%%%%%%%%%%%%%%%%%%%%%%%%%%%%%%%%%%%%%%%%%%%%%%%%%%%%%%%%%%%%%%%%%%%

\printbibliography

\end{document}